\documentclass[preprint]{article}

\usepackage[position]{neurips_2025}

\usepackage[utf8]{inputenc} 
\usepackage[T1]{fontenc}    
\usepackage{hyperref}       
\usepackage{url}            
\usepackage{booktabs}       
\usepackage{amsfonts}       
\usepackage{nicefrac}       
\usepackage{microtype}      
\usepackage{xcolor}         
\usepackage{xspace}
\usepackage{amsmath}
\usepackage{graphicx}

\newcommand{\workname}{\texttt{Collaborative Thoughts}\xspace}

\title{Reasoning with Autoregressive-Diffusion\\Collaborative Thoughts}

\author{%
    Mu Yuan$^{1}$\thanks{Mu Yuan and Liekang Zeng contributed equally to this work.},\ 
    ~~Liekang Zeng$^{1}$\footnotemark[1],\
    ~~Guoliang Xing$^{1}$,\
    ~~Lan Zhang$^{2}$,\
    ~~Yunhao Liu$^{3}$\\
  $^{1}$The Chinese University of Hong Kong\\
  $^{2}$University of Science and Technology of China\\
  $^{3}$Tsinghua University\\
  \texttt{muyuan@cuhk.edu.hk, lkzeng@cuhk.edu.hk, glxing@cuhk.edu.hk}\\ 
  \texttt{zhanglan@ustc.edu.cn, yunhao@tsinghua.edu.cn} \\
}

\begin{document}

\maketitle

\begin{abstract}
Autoregressive and diffusion models represent two complementary generative paradigms.
Autoregressive models excel at sequential planning and constraint composition, yet struggle with tasks that require explicit spatial or physical grounding.
Diffusion models, in contrast, capture rich spatial structure through high-dimensional generation, but lack the stepwise logical control needed to satisfy complex, multi-stage constraints or to reliably identify and correct errors.

We introduce \workname, a unified collaborative framework that enables autoregressive and diffusion models to reason and generate jointly through a closed-loop interaction.
In \workname, autoregressive models perform structured planning and constraint management, diffusion models instantiate these constraints as intermediate visual thoughts, and a vision-based critic module evaluates whether the visual thoughts satisfy the intended structural and physical requirements.
This feedback is then used to iteratively refine subsequent planning and generation steps, mitigating error propagation across modalities.
Importantly, \workname uses the same collaborative loop regardless of whether the task is autoregressive question answering or diffusion-based visual generation.
Through representative examples, we illustrate how \workname can improve the reliability of spatial reasoning and the controllability of generation.
\end{abstract}

\section{Introduction}

\begin{figure}[t]
    \centering
    \includegraphics[width=0.95\linewidth]{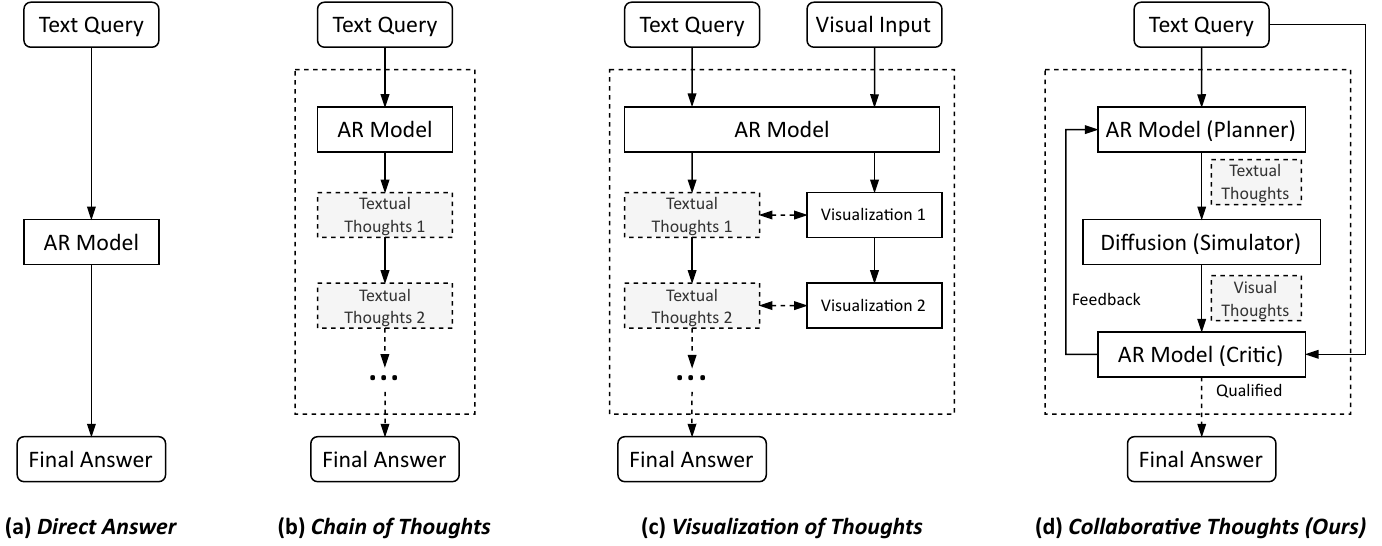}
    \caption{Traditional chain-of-thought (CoT) \cite{wei2022chain} ponders queries via only text, and Visualization of Thoughts (VoT) \cite{wu2024mind,li2025imagine} relies on visual input to initiate the visualization of thinking traces. \workname orchestrates autoregressive and diffusion models to collaboratively think via multimodal reasoning traces.
    }
    \vspace{-0.2in}
    \label{fig:overview}
\end{figure}

Reasoning constitutes the fundamental capability of general-purpose artificial intelligence, enabling systems to decompose complex problems, formulate structured plans, and execute multi-step solutions.
Beyond direct answer (Figure \ref{fig:overview}a), the Chain-of-Thought (CoT) paradigm~\cite{wei2022chain} has significantly advanced this frontier, empowering Large Language Models (LLMs) to tackle intricate tasks through intermediate verbal reasoning (Figure \ref{fig:overview}b).
While effective for semantic deduction, this reliance on symbolic abstraction creates a "blind spot" for tasks requiring explicit spatial awareness or physical common sense: LLMs often struggle to verify geometric structures or simulate physical interactions, leading to hallucinations that are linguistically coherent but physically implausible~\cite{chen2024spatialvlm,ranasinghe2024learning,wu2024mind}.

To bridge this disconnect, emerging research has pivoted toward visualizing text thoughts~\cite{li2025imagine,wu2024mind}, a paradigm that seeks to externalize reasoning through visual formats (Figure \ref{fig:overview}c).
These approaches augment LLMs with the capacity to generate schematic diagrams or intermediate imagery to serve as cognitive scaffolding.
However, in these approaches, generated visual contents are typically treated as immutable ground truth rather than provisional hypotheses.
Without an explicit feedback mechanism, errors introduced during visual generation cannot be effectively detected or corrected, leading to irreversible error propagation.
Moreover, most prior methods focus on annotating or interpreting static visual inputs, leaving the potential of an iterative, self-correcting interaction between reasoning and generation largely unexplored.

To address these limitations, we explore to synergizing two complementary yet fundamentally different generative paradigms: \textit{autoregressive models} and \textit{diffusion models}.
Autoregressive (AR) models, exemplified by LLMs, excel at symbolic composition and sequential constraint management but struggle to faithfully represent high-dimensional geometric structures (Figure~\ref{fig:demo-correctness}, AR-Only). Diffusion models, in contrast, define the state-of-the-art in high-dimensional visual synthesis and act as powerful generators of spatially coherent visual content~\cite{croitoru2023diffusion,yang2023diffusion}.
Combining the both worlds, we propose \workname, a closed-loop framework that enables sustained collaboration between autoregressive and diffusion paradigms through iterative planning, generation, and refinement.
This perspective is inspired by Dual Coding Theory~\cite{paivio1991dual}, which posits that human cognition integrates sequential symbolic reasoning with spatial mental imagery.
When solving complex geometric or physical problems, humans repeatedly construct mental images, evaluate them against physical constraints, and revise their reasoning accordingly.
Motivated by this cognitive process, rather than viewing generation as a one-shot process, our framework treats intermediate outputs as hypotheses that can be inspected and refined.

Concretely, as shown in Figure \ref{fig:overview}d, autoregressive models are responsible for structured planning and constraint composition, decomposing tasks into a sequence of visualizable requirements.
Diffusion models instantiate these requirements as intermediate visual blueprints that explicitly capture spatial structure.
A critic module then evaluates the generated visual thoughts against the intended constraints and provides feedback that guides subsequent planning and generation steps.
This iterative \emph{Simulate-Critic-Refine} cycle mitigates error propagation from modality misalignment and enables continuous correction across reasoning and generation.

Importantly, \workname is agnostic to the task's final output modality.
For tasks that require symbolic or textual answers, diffusion models primarily serve to generate intermediate visual references that ground autoregressive reasoning.
For tasks that require visual generation, autoregressive models assist diffusion by providing stepwise planning and refinement, while the diffusion model produces the final output.
In both cases, the same closed-loop interaction governs simulate, critic, and refinement, offering a unified treatment of reasoning and generation across paradigms.

As a proof of concept, in this work, we focus on representative examples that illustrate the capabilities and opportunities enabled by this collaborative framework.
These examples demonstrate how autoregressive-diffusion collaboration can improve the reliability of spatial reasoning and the controllability of generative processes, pointing toward a broader research direction on collaborative reasoning across generative paradigms.

\section{Related Work}

\textbf{Autoregressive-Guided Visual Planning and Layout Generation.}
A line of work integrates autoregressive models into text-to-image generation pipelines to alleviate the limitations of fixed text encoders by introducing intermediate planning or layout representations.
Representative approaches such as LLM-grounded Diffusion~\cite{lian2024llm} and LayoutLLM~\cite{qu2023layoutllm} decompose user prompts into explicit spatial layouts or bounding boxes that condition subsequent diffusion generation.
RPG~\cite{yang2024mastering} further extends this idea by using multimodal autoregressive models to perform multi-step planning for region-level generation, enabling better handling of compositional attributes.
DiffusionGPT~\cite{qin2024diffusiongpt} explores routing prompts through structured reasoning paths before generation.
Despite their effectiveness, these approaches are predominantly sequential: planning is performed once, and the generated visual output is treated as a fixed realization of that plan.
In contrast, our framework emphasizes interaction, where intermediate visual outputs are explicitly checked and used to refine subsequent planning steps.

\textbf{Iterative Visual Refinement with Feedback.}
To mitigate errors in generative models, several studies introduce iterative refinement mechanisms based on visual feedback.
Self-correcting diffusion pipelines use autoregressive detectors to compare generated images against textual requirements and issue corrective signals for subsequent generations.
More general frameworks, such as Iterative Prompt Refinement~\cite{jeon2025iterative}, leverage vision-language models to adjust prompts based on discrepancies observed in generated visuals.
Related work on visual reasoning, including Visual Sketchpad and Visual Chain-of-Thought~\cite{hu2024visual}, shows that intermediate visual artifacts can support geometric and logical inference.
While effective at reducing generation errors, these methods typically treat visual outputs as auxiliary signals for validation.
Our approach differs in that visual generation is integrated as a central intermediate state in the reasoning process, enabling planning decisions to be directly informed by verified visual structure rather than post hoc correction alone.

\textbf{Simulation-Based Reasoning and Physical Grounding.}
Simulation has also been explored as a means to ground reasoning in physical dynamics.
The Mind's Eye framework~\cite{liu2023minds} employs external physics engines, such as MuJoCo~\cite{todorov2012mujoco}, to simulate outcomes for physical reasoning tasks.
Recent advances in video generation have inspired approaches like PhysGen~\cite{liu2024physgen} and InterDyn~\cite{akkerman2025interdyn}, which use diffusion-based video models to predict object dynamics and interactions.
In robotics, methods such as CoT-VLA~\cite{zhao2025cot} and Vidarc~\cite{feng2025vidarc} leverage imagined future visual states to guide control policies.
However, these approaches often rely on a single simulation modality and lack mechanisms for structured, stepwise verification across reasoning and generation.
Our work complements this line of research by focusing on collaborative interaction between autoregressive planning and diffusion-based visual generation, enabling iterative verification and refinement without assuming access to precise physical simulators.

\section{Method}

\workname is a framework that synergizes autoregressive reasoning with diffusion-based visual simulation, mainly comprising three main components (Figure \ref{fig:overview}d).
Unlike rigorous open-loop generation, our approach treats visual synthesis as an iterative optimization process guided by semantic constraints.

\subsection{Problem Formulation}
Let $\mathcal{Q}$ be a natural language query requiring spatial or physical reasoning, and $\mathcal{A}$ be the target answer.
We model the reasoning process as a sequential decision-making problem over $T$ steps.
At each step $t$, the system maintains a reasoning state $\mathcal{S}_t = \{P_t, R_t, F_t\}$, consisting of a textual thought $P_t$ (i.e., for visual prompt), a generated visual thought $R_t$ (e.g., image), and textual feedback $F_t$.
Assuming the knowledge of the physical world aligns with the distribution $p_{\text{world}}$, our goal is to maximize the joint probability of the correct answer $\mathcal{A}$ with respect to $p_{\text{world}}$ given the query and the evolved visual thoughts:
\begin{align}
\mathcal{A}^* = \arg\max_{\mathcal{A}} \mathcal{M} (\mathcal{A} \mid \mathcal{Q}, R^*) \sim p_{\text{world}},    
\end{align}
where $R^*$ is the optimal visual simulation selected from the iterative trajectory.
The framework comprises three coupled agents: the Planner (autoregressive model), the Simulator (diffusion model), and the Critic (autoregressive model).

\subsection{The Planner: Semantic-to-Visual Translation}
The Planner, denoted as $\mathcal{M}_{plan}$, is an autoregressive LLM responsible for reasoning decomposition and prompt engineering.
In the initial step ($t=0$), the Planner analyzes $\mathcal{Q}$ to extract implicit spatial constraints (e.g., "object A must support object B") and generates an initial scene description prompt $P_0$.
In subsequent steps ($t>0$), the Planner acts as a refiner.
It receives the feedback $F_{t-1}$ from the Critic, which details the discrepancies between the previous simulation $I_{t-1}$ and the physical constraints.
The Planner then updates the prompt to rectify these errors:
\begin{align}
P_t = \mathcal{M}_{\text{plan}}(\mathcal{Q}, F_{t-1}, H_{t-1}),
\label{eq:planner}
\end{align}
where $H_{t-1}$ represents the conversation history, enabling the model to perform in-context learning from past failures.

\subsection{The Simulator: Physical Instantiation}
The Simulator $\mathcal{M}_{\text{sim}}$ serves as the external "world model".
It maps the semantic instructions $P_t$ into a pixel-space representation $R_t$.
To mitigate the stochastic instability of diffusion models, layout-guided generation strategies (e.g., ControlNet \cite{zhang2023adding}) may be employed.
This ensures that the global spatial structure adheres to the Planner's intent while allowing the diffusion model to fill in fine-grained physical details (texture, lighting, occlusion):
\begin{align}
  R_t = \mathcal{M}_{\text{sim}}(P_t | \mathcal{C}) \sim p_{\text{world}}.
  \label{eq:simulator}
\end{align}
Here, $\mathcal{C}$ denotes optional structural constraints (such as bounding boxes or depth maps) generated by the Planner to enforce geometric stability.
The generation of the diffusion model $\mathcal{M}_{\text{sim}}$ conveys its physical knowledge to $R_t$ by approximating the distribution of the real world $p_{\text{world}}$.

\subsection{The Critic: Visual-Logic Alignment}
The Critic, $\mathcal{M}_{\text{critic}}$, is an autoregressive model (e.g., VLM) tasked with supervision.
It closes the reasoning loop by perceiving the generated visual thought $R_t$ and comparing it against the original query requirements.
As shown in Equation \eqref{eq:critic}, the Critic performs two functions: 
1) Verification Score ($v_t$): A scalar score $v_t \in [0, 1]$ indicating the degree of constraint satisfaction (e.g., checking for object hallucinations, floating objects, or incorrect relative positions);
2) Corrective Feedback ($F_t$): If $v_t$ falls below a confidence threshold $\tau$, the Critic generates natural language feedback describing the specific violation (e.g., "The red cube is floating; it must rest on the table surface").
\begin{align}
v_t, F_t = \mathcal{M}_{\text{critic}}(R_t, \mathcal{Q}).
\label{eq:critic}    
\end{align}

\subsection{Termination and Inference}
The iterative process terminates when one of the following conditions is met:
1) Convergence: The verification score exceeds the threshold ($v_t > \tau$);
2) Budget Exhaustion: The iteration count reaches the maximum limit ($t \ge T_{max}$);
3) Deadlock: Semantic oscillation is detected in consecutive prompts.

Upon termination, the critic synthesizes the final answer $\mathcal{A}$ based on the validated visual thought $R^*$ (or the highest-scoring reference $R_{\text{best}}$ in the trajectory), effectively grounding the verbal reasoning in verified visual simulation.

\section{Demonstrations}
\label{sec:demonstrations}

\begin{figure}[t]
  \centering
  \includegraphics[width=1\linewidth]{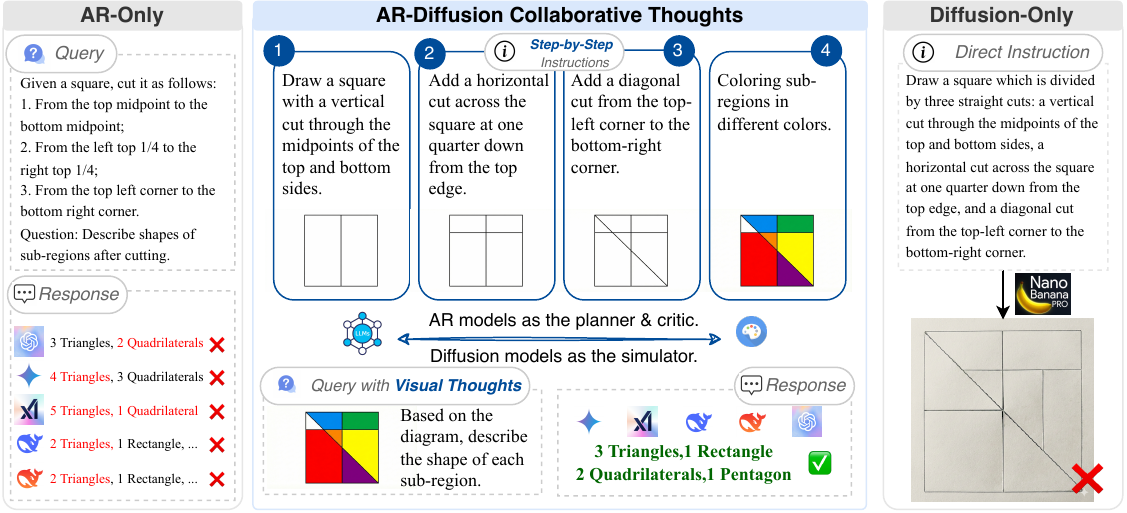}
  \caption{
  While text-based reasoning (AR-Only) struggles with spatial abstraction and direct generation (Diffusion-Only) lacks structural logic, our framework leverages the complementary strengths of both. The autoregressive (AR) model acts as a planner to break down the diffusion process into four discrete steps. This step-by-step visual verification prevents the error propagation seen in the baselines, allowing for accurate identification of the final geometric shapes.
  }
  \label{fig:demo-correctness}
\end{figure}

Figure \ref{fig:demo-correctness} qualitatively demonstrates the superiority of the proposed AR-Diffusion collaborative thinking framework over uni-architecture approaches in handling complex spatial reasoning tasks.
The demonstration involves a multi-step geometric problem requiring the sequential cutting of a square and the subsequent identification of the resulting sub-regions.
As illustrated in the "AR-Only" panel, purely text-based autoregressive models struggle with spatial abstraction; despite receiving clear instructions, they hallucinate incorrect geometric compositions (e.g., miscounting triangles and quadrilaterals) due to a lack of visual grounding.
Conversely, the "Diffusion-Only" approach (right panel) attempts to generate the visual solution in a single pass.
However, it fails to adhere to the strict structural logic required, resulting in imprecise cuts and geometric distortions that render the output unusable for accurate reasoning.
In contrast, the central panel highlights the efficacy of our collaborative framework.
By leveraging the AR model as a high-level planner and the diffusion model as an external visualizer, the system decomposes the query into a discrete chain of visual steps—ranging from the initial vertical cut to the final coloring of sub-regions.
This step-by-step visual refinement process creates a precise "blueprint," effectively preventing error propagation.
Consequently, the system correctly identifies the complex final configuration (3 triangles, 1 rectangle, 2 quadrilaterals, and 1 pentagon), validating the necessity of interleaving reasoning and generation for spatial tasks.

\begin{figure}[t]
  \centering
  \includegraphics[width=1\linewidth]{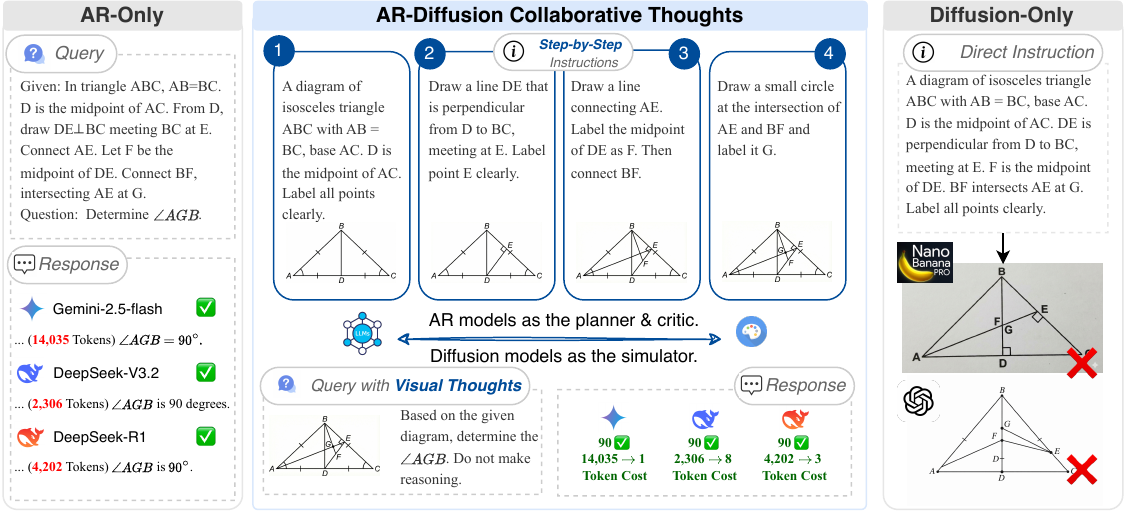}
  \caption{Illustration of reasoning paradigms for geometric problem solving. The proposed AR-Diffusion \workname (center) creates intermediate visual blueprints to bridge the gap between text and vision. This approach corrects the geometric hallucinations observed in Diffusion-Only methods (right) and significantly improves inference efficiency compared to the AR-Only textual chain-of-thought (left), reducing computational costs by orders of magnitude.}
  \label{fig:demo-few-token}
\end{figure}

Complementing the topological decomposition task in Figure \ref{fig:demo-correctness}, Figure \ref{fig:demo-few-token} further illustrates the versatility of the \workname framework in the context of Euclidean geometry problem solving.
This demonstration specifically highlights the framework's ability to bridge the gap between accurate visual generation and efficient logical inference. 
As depicted in the "AR-Only" panel (left), while advanced Large Language Models (e.g., Gemini-1.5-Pro, DeepSeek-R1) can deductively solve for the target angle $\angle AGB$, the process is computationally expensive. 
Relying solely on textual Chain-of-Thought requires parsing complex geometric relationships without visual grounding, resulting in excessive token consumption (up to 14,035 tokens) to reach the correct conclusion. Conversely, the "Diffusion-Only" baseline (right) attempts to generate the diagram via direct instruction but suffers from "geometric hallucination," failing to preserve essential constraints such as perpendicularity ($DE \perp BC$) or precise intersections, rendering the image useless for reasoning. The central panel demonstrates how the proposed collaborative framework resolves these limitations.
By utilizing the autoregressive model to plan a sequential construction, the system guides the diffusion model to generate a geometrically rigorous intermediate diagram. This high-fidelity visualization allows the reasoning model to bypass lengthy textual derivations, grounding its answer directly in the visual data. The result is a dramatic optimization in inference efficiency: the reasoning cost is reduced by orders of magnitude (e.g., from 14,035 tokens to a single token) while maintaining $100\%$ accuracy.
Together, Figures 1 and 2 validate that \workname not only prevents error propagation in complex spatial tasks but also provides a scalable, low-latency paradigm for geometric reasoning.

\section{Discussion}

\textbf{Why does \workname work?}
The efficacy of our framework partially stems from leveraging the diffusion model as a data-driven implicit world model that complements the abstract logic of LLMs.
Having internalized vast phenomenological knowledge (e.g., occlusion, gravity, material properties) from large-scale pre-training, the diffusion model acts as a 'Physical Consistency Filter', forcing the planner's symbolic hypotheses to confront the statistical realities of pixel space.
By grounding textual reasoning in these robust visual priors, the system effectively bridges the gap between semantic coherence and physical viability, rejecting logically sound but physically implausible solutions.

\textbf{The "Soft Simulator" Paradigm.}
\workname empirically grounds Dual Coding Theory\cite{paivio1991dual} in computational systems by mutually coupling autoregressive LLMs and diffusion models in an iterative loop.
Unlike rigid physics engines (e.g., MuJoCo \cite{todorov2012mujoco}) that require precise parameterization, our framework establishes diffusion models as "soft simulators", which also suffer from the "sim-to-real" gap.
This trades absolute numerical precision for semantic universality, enabling physical reasoning in open-domain scenarios (e.g., "stacking heterogeneous household objects") where formal modeling is intractable.

\textbf{Simulator and Critic Bottlenecks.} 
The iterative nature of the "Simulate-Critic-Refine" cycle introduces significant computational overhead from the diffusion-based visual generation.
Also, the reasoning upper bound is constrained by the diffusion model's physical understanding and generation capabilities, as well as the critic's verification correctness.

\section{Conclusion}

We presented \workname, a framework that bridges the chasm between autoregressive semantic planning and diffusion-based visual simulation. By treating visual simulation as an iterative reasoning step rather than a final output, we demonstrated that a closed "Generation-Reasoning" loop effectively grounds abstract logic in pixel-level physical priors, mitigating the spatial hallucinations common in pure LLMs.
Our findings substantiate a broader paradigm shift: robust spatial intelligence arises not from a single omnipotent model, but from a synergistic cognitive architecture where the logical rigor of CoT and the phenomenological intuition of diffusion-based simulation continuously critique and refine one another.

For future work, we explore to address the inference overhead of the iterative cycle to accelerate the feedback loop. Furthermore, we aim to extend the "simulator" from 2D static visual content to 3D assets and video dynamics, paving the way for embodied agents that can mentally rehearse physical actions before real-world execution.

{\small
\bibliographystyle{abbrv}
\bibliography{ref}

@article{lian2024llm,
  title={LLM-grounded Diffusion: Enhancing Prompt Understanding of Text-to-Image Diffusion Models with Large Language Models},
  author={Lian, Long and Li, Boyi and Yala, Adam and Darrell, Trevor},
  journal={Transactions on Machine Learning Research},
  year={2024}
}

@inproceedings{qu2023layoutllm,
  title={LayoutLLM-T2I: Eliciting Layout Guidance from LLM for Text-to-Image Generation},
  author={Qu, Leigang and Wu, Shengqiong and Fei, Hao and Nie, Liqiang and Chua, Tat-Seng},
  booktitle={Proceedings of the 31st ACM International Conference on Multimedia},
  year={2023}
}

@inproceedings{yang2024mastering,
  title={Mastering Text-to-Image Diffusion: Recaptioning, Planning, and Generating with Multimodal LLMs},
  author={Yang, Ling and Yu, Zhaochen and Meng, Chenlin and Xu, Minkai and Ermon, Stefano and Cui, Bin},
  booktitle={International Conference on Machine Learning (ICML)},
  year={2024}
}

@article{qin2024diffusiongpt,
  title={DiffusionGPT: LLM-Driven Text-to-Image Generation System},
  author={Qin, Jie and Wu, Jie and Chen, Weifeng and Ren, Yuxi and Li, Huixia and Wu, Hefeng and Xiao, Xuefeng and Wang, Rui and Wen, Shilei},
  journal={arXiv preprint arXiv:2401.10061},
  year={2024}
}

@inproceedings{jeon2025iterative,
  title={Iterative Prompt Refinement for Safer Text-to-Image Generation},
  author={Jeon, Jinwoo and Oh, JunHyeok and Lee, Hayeong and Lee, Byung-Jun},
  booktitle={Proceedings of the 2025 Conference on Empirical Methods in Natural Language Processing (EMNLP)},
  year={2025}
}

@inproceedings{todorov2012mujoco,
  title={Mujoco: A physics engine for model-based control},
  author={Todorov, Emanuel and Erez, Tom and Tassa, Yuval},
  booktitle={2012 IEEE/RSJ international conference on intelligent robots and systems},
  pages={5026--5033},
  year={2012},
  organization={IEEE}
}

@inproceedings{akkerman2025interdyn,
  title={InterDyn: Controllable interactive dynamics with video diffusion models},
  author={Akkerman, Rick and Feng, Haiwen and Black, Michael J and Tzionas, Dimitrios and Abrevaya, Victoria Fern{\'a}ndez},
  booktitle={Proceedings of the Computer Vision and Pattern Recognition Conference},
  pages={12467--12479},
  year={2025}
}

@article{wei2022chain,
  title={Chain-of-thought prompting elicits reasoning in large language models},
  author={Wei, Jason and Wang, Xuezhi and Schuurmans, Dale and Bosma, Maarten and Xia, Fei and Chi, Ed and Le, Quoc V and Zhou, Denny and others},
  journal={Advances in neural information processing systems},
  volume={35},
  pages={24824--24837},
  year={2022}
}

@inproceedings{hu2024visual,
  title={Visual Sketchpad: Sketching as a Visual Chain of Thought for Multimodal Language Models},
  author={Hu, Yushi and Shi, Weijia and Fu, Xingyu and Roth, Dan and Ostendorf, Mari and Zettlemoyer, Luke and Smith, Noah A and Krishna, Ranjay},
  booktitle={Advances in Neural Information Processing Systems (NeurIPS)},
  year={2024}
}

@inproceedings{liu2023minds,
  title={Mind's Eye: Grounded Language Model Reasoning through Simulation},
  author={Liu, Ruibo and Wei, Jason and Gu, Shixiang Shane and Wu, Te-Yen and Vosoughi, Soroush and Cui, Claire and Zhou, Denny and Dai, Andrew M},
  booktitle={International Conference on Learning Representations (ICLR)},
  year={2023}
}

@inproceedings{liu2024physgen,
  title={PhysGen: Rigid-Body Physics-Grounded Image-to-Video Generation},
  author={Liu, Shaowei and Zhang, Zhongzheng and Zhang, Jason and Xu, Danfei and Zhu, Jun-Yan},
  booktitle={European Conference on Computer Vision (ECCV)},
  year={2024}
}

@inproceedings{zhao2025cot,
  title={CoT-VLA: Visual Chain-of-Thought Reasoning for Vision-Language-Action Models},
  author={Zhao, Qingqing and Lu, Yao and Kim, Moo Jin and Fu, Zipeng and Zhang, Zhuoyang and Wu, Yecheng and Li, Max and Ma, Qianli and Han, Song and Finn, Chelsea and Handa, Ankur and Liu, Ming-Yu and Xiang, Donglai and Wetzstein, Gordon and Lin, Tsung-Yi},
  booktitle={Proceedings of the IEEE/CVF Conference on Computer Vision and Pattern Recognition (CVPR)},
  year={2025}
}

@article{feng2025vidarc,
  title={Vidarc: Embodied Video Diffusion Model for Closed-loop Control},
  author={Feng, Yao and Xiang, Chendong and Mao, Xinyi and Tan, Hengkai and Zhang, Zuyue and Huang, Shuhe and Zheng, Kaiwen and Liu, Haitian and Su, Hang and Zhu, Jun},
  journal={arXiv preprint arXiv:2512.17661},
  year={2025}
}

@inproceedings{chen2024spatialvlm,
  title={Spatialvlm: Endowing vision-language models with spatial reasoning capabilities},
  author={Chen, Boyuan and Xu, Zhuo and Kirmani, Sean and Ichter, Brain and Sadigh, Dorsa and Guibas, Leonidas and Xia, Fei},
  booktitle={Proceedings of the IEEE/CVF Conference on Computer Vision and Pattern Recognition},
  pages={14455--14465},
  year={2024}
}

@inproceedings{ranasinghe2024learning,
  title={Learning to localize objects improves spatial reasoning in visual-llms},
  author={Ranasinghe, Kanchana and Shukla, Satya Narayan and Poursaeed, Omid and Ryoo, Michael S and Lin, Tsung-Yu},
  booktitle={Proceedings of the IEEE/CVF Conference on Computer Vision and Pattern Recognition},
  pages={12977--12987},
  year={2024}
}

@article{wu2024mind,
  title={Mind's eye of LLMs: visualization-of-thought elicits spatial reasoning in large language models},
  author={Wu, Wenshan and Mao, Shaoguang and Zhang, Yadong and Xia, Yan and Dong, Li and Cui, Lei and Wei, Furu},
  journal={Advances in Neural Information Processing Systems},
  volume={37},
  pages={90277--90317},
  year={2024}
}

@article{croitoru2023diffusion,
  title={Diffusion models in vision: A survey},
  author={Croitoru, Florinel-Alin and Hondru, Vlad and Ionescu, Radu Tudor and Shah, Mubarak},
  journal={IEEE transactions on pattern analysis and machine intelligence},
  volume={45},
  number={9},
  pages={10850--10869},
  year={2023},
  publisher={Ieee}
}

@article{yang2023diffusion,
  title={Diffusion models: A comprehensive survey of methods and applications},
  author={Yang, Ling and Zhang, Zhilong and Song, Yang and Hong, Shenda and Xu, Runsheng and Zhao, Yue and Zhang, Wentao and Cui, Bin and Yang, Ming-Hsuan},
  journal={ACM computing surveys},
  volume={56},
  number={4},
  pages={1--39},
  year={2023},
  publisher={ACM New York, NY, USA}
}

@inproceedings{li2025imagine,
  title={Imagine While Reasoning in Space: Multimodal Visualization-of-Thought},
  author={Li, Chengzu and Wu, Wenshan and Zhang, Huanyu and Xia, Yan and Mao, Shaoguang and Dong, Li and Vuli{\'c}, Ivan and Wei, Furu},
  booktitle={Forty-second International Conference on Machine Learning}
}

@article{paivio1991dual,
  title={Dual coding theory: Retrospect and current status.},
  author={Paivio, Allan},
  journal={Canadian Journal of Psychology/Revue canadienne de psychologie},
  volume={45},
  number={3},
  pages={255},
  year={1991},
  publisher={Canadian Psychological Association}
}

@inproceedings{zhang2023adding,
  title={Adding conditional control to text-to-image diffusion models},
  author={Zhang, Lvmin and Rao, Anyi and Agrawala, Maneesh},
  booktitle={Proceedings of the IEEE/CVF international conference on computer vision},
  pages={3836--3847},
  year={2023}
}
}

\end{document}